\pdfoutput=1
\documentclass[11pt]{article}

\usepackage{arabtex}
\usepackage{utf8}

\usepackage{acl}

\usepackage{times}
\usepackage{latexsym}

\usepackage[T1]{fontenc}

\usepackage[utf8]{inputenc}

\usepackage{microtype}

\usepackage{inconsolata}

\setcode{utf8}
\usepackage{graphicx}
\usepackage{algorithm}
\usepackage{algpseudocode}
\usepackage{amsmath}
\usepackage{multirow}

\title{AraS2P: Arabic Speech-to-Phonemes System}

\author{
  Bassam Mattar \\
  Alexandria University \\
  \texttt{b.mattar@alexu.edu.eg} \\
  \And
  Mohamed Fayed \\
  Applied innovation Center \\
  m.essam@aic.gov.eg \\
  Georgia Institute of Technology \\
  \texttt{mfayed8@gatech.edu} \\
  \And
  Ayman Khalafallah \\
   Applied innovation Center \\
  a.khalafallah@aic.gov.eg
}

\begin{document}
\maketitle

\begin{abstract}
This paper describes AraS2P, our speech-to-phonemes system submitted to the Iqra’Eval 2025 Shared Task.
We adapted Wav2Vec2-BERT via Two-Stage training strategy.
In the first stage, task-adaptive continue pretraining was performed on large-scale Arabic speech-phonemes datasets, which were generated by converting the Arabic text using the MSA Phonetiser.
In the second stage, the model was fine-tuned on the official shared task data, with additional augmentation from XTTS-v2-synthesized recitations featuring varied Ayat segments, speaker embeddings, and textual perturbations to simulate possible human errors.
The system ranked first on the official leaderboard, demonstrating that phoneme-aware pretraining combined with targeted augmentation yields strong performance in phoneme-level mispronunciation detection.
\end{abstract}

\section{Introduction}

Automatic mispronunciation detection and diagnosis (MDD) plays a key role in computer-aided pronunciation learning (CAPL), providing learners with objective and scalable feedback on their pronunciation quality score~\cite{kheir2023automatic}.
In Arabic, MDD is particularly challenging due to the language’s complex phonemic inventory, the presence of emphatic and pharyngeal consonants, and the semantic role of short vowels (diacritics) \cite{abdou2014}.
These characteristics make accurate phoneme-level detection especially important, as even subtle deviations can significantly change meaning. 



In this work, we describe a system based on a Wav2Vec2-BERT architecture \cite{baevski2020} that employs a two-stage training strategy:
(1) task-adaptive continue pretraining on large Arabic speech datasets—Common Voice (Arabic split), SADA, and MASC—using phoneme-level supervision generated via the MSA Phonetiser,\footnote{\url{https://github.com/Iqra-Eval/MSA_phonetiser}} resulting in labeled corpora that capture fine-grained phonetic distinctions, and
(2) fine-tuning on the official shared task data as well as targeted augmentation through XTTS-v2-synthesized recitations that vary in Ayat segments, speaker embeddings, and noisy textual content to simulate realistic recitations errors. 

We summarize our contributions as follows:
\begin{itemize}
  \item A phoneme-aware task-adaptive pretraining strategy for Arabic MDD using large-scale speech-phonemes data.
  \item A targeted augmentation pipeline where we add noise to text, convert the noisy text to phonemes using MSA-Phonetizer,  and generate corresponding speech for many speakers using XTTS-v2~\cite{casanova2024xtts}.
  \item Our model ranks first on the Iqra’Eval 2025 benchmark leaderboard, demonstrating effectiveness of our training strategy.
\end{itemize}

\section{Related Work}

\subsection{Arabic CAPL and Mispronunciation Detection}

Computer-Assisted Pronunciation Learning (CAPL) systems rely on Mispronunciation Detection and Diagnosis (MDD) to provide automated feedback for learners \cite{witt2000, eskenazi2009}.
Early MDD approaches often derived pronunciation quality metrics from acoustic likelihoods computed from recognition results, such as the Goodness of Pronunciation (GOP) score \cite{witt2000}.
While GOP provides a practical way to detect pronunciation deviations, its granularity is limited to the phone level and its accuracy can be affected by recognition errors.
Other research \cite{bonaventura2000phonetic, raux2002automatic} has enhanced pronunciation modeling by incorporating likely pronunciation variants into a pronunciation dictionary, which can involve manual specification of error patterns.

Recent years have seen the adoption of deep learning and end-to-end architectures for MDD, enabling systems to learn pronunciation error patterns directly from data \cite{peng2022text}.
For Arabic, MDD poses additional challenges due to its rich consonant inventory, emphatic and pharyngeal sounds, and the omission of short vowels in most written text and ASR systems \cite{elkheir2025}.
Consequently, slight pronunciation errors—such as mixing up emphatic and non-emphatic consonants—may change the meaning of a word.

Arabic MDD research has explored handcrafted acoustic features, CNN-based classifiers, and transfer learning from large-scale ASR models \cite{calik2023, alrashoudi2025improving}.
Several works have focused on Qur’anic recitation, where precise phoneme articulation is central \cite{abdou2014, alrumiah2023intelligent, harere2023quran}.
\cite{elkheir2025} provided the first publicly available benchmark for Arabic phoneme-level MDD, using Qur’anic recitation with time-aligned phoneme annotations. 

\subsection{Self-Supervised Phoneme Recognition Models}

Self-supervised learning has significantly advanced phoneme recognition, which in turn has improved the performance of MDD systems.
Wav2Vec-BERT 2.0 model \cite{baevski2020} learns contextualized speech representations from raw audio by combining a convolutional encoder with a Transformer context network~\cite{devlin2019bert, baevski2019vq}.
It was pretrained using a contrastive objective~\cite{chen2020simple, he2020momentum} over masked audio segments, then fine-tuned with a Connectionist Temporal Classification (CTC) objective~\cite{graves2006connectionist}.
Wav2vec 2.0 achieves state-of-the-art performance in phoneme recognition tasks, making it well-suited for MDD. 

Building on this, Wav2Vec-BERT integrates a BERT-style masked language modeling (MLM) objective~\cite{devlin2019bert}  with the Wav2Vec 2.0 framework \cite{chung2021w2vbert}.
This joint optimization learns both quantized acoustic units and contextual relationships between them, producing richer and more discriminative phonetic representations.
Instead of iteratively re-clustering discrete units like HuBERT~\cite{hsu2021}, w2v-BERT learns quantization and context modeling in a single end-to-end process.

Multilingual Wav2Vec-BERT 2.0 extends this approach to $143$ languages using over $4.5$ million hours of speech for pretraining \cite{barrault2023seamless}.
Its large-scale multilingual exposure enables robust representation of fine phonetic distinctions, even in low-resource settings like Arabic MDD.
Compared to Wav2Vec 2.0, Wav2Vec-BERT 2.0 incorporates MLM-based contextual modeling directly into the acoustic encoder, allowing it to learn longer-range phoneme patterns.
For this reason, we used Wav2Vec-BERT 2.0 pretrained weights.

\subsection{Benchmarks and Shared Tasks}

Iqra’Eval Shared Task \cite{elkheir2025} represents a milestone for Arabic MDD by offering a publicly available benchmark, standardized evaluation protocol, and a leaderboard for reproducible comparison.
Similar to MGB Challenge for Arabic ASR \cite{ali2016mgb} and other shared tasks in speech and Natural Language Processing (NLP), this benchmark has stimulated community engagement and methodological innovation.
Through integrating controlled evaluation with phoneme-level detection, Iqra’Eval addresses a critical gap in Arabic CAPL research by establishing a standardized benchmark for systematic evaluation. 

\section{Two-Stage Training}
We adapted Wav2Vec2-BERT~\cite{barrault2023seamless} to our downstream task via Two-Stage training.
We continued pretraining it on Arabic speech-phonems pairs (section~\ref{ssect:acpt}).
Meanwhile, we conducted exploratory data analysis to measure the alignment between continue pretraining and fine-tuning (section~\ref{ssect:eda}).
Finally, we utilized training set of the task as well as our synthetically generated dataset for fine-tuning (section~\ref{ssect:ft}).

\subsection{Adaptive Continue Pretraining}\label{ssect:acpt}
Continue pretraining has shown to be an effective technique to improve the performance of pretrained models on languages of interest~\cite{kalyan2021ammus, zhou2024continual, fujii2024continual,  alves2024tower}.
To boost our model, we continued pretraining it on speech-phonemes pairs.
We deployed MSA-phonetizer\footnote{\url{https://github.com/Iqra-Eval/MSA_phonetiser}} to convert open-source datasets with speech-text pairs into speech-phonemes pairs, hence adapting it to suite the downstream task (Adaptive Continue Pretraining).
Specifically, our pretraining data is constructed from Common Voice Arabic split~\cite{ardila2019common}, SADA\cite{alharbi2024sada} and MASC\cite{masc} datasets.
Table\ref{tab:data-stats} includes statistics about these datasets.
\begin{table}[ht]
\centering
\begin{tabular}{lc}
Dataset & size (hours) \\
\hline
Common Voice (Ar-Split) & 157 \\
SADA & 668 \\
MASC & 1,000 \\
\end{tabular}
\caption{Statistics of datasets used in our adaptive continue pretraining stage.}
\label{tab:data-stats}
\end{table}

We used Adam optimizer with weight decay~\cite{loshchilov2017decoupled}.
We set hyper-parameters as follows: learning rate of $1 \times 10^{-5}$, Linear Decay scheduler, weight decay equals to $0.01$, Adam betas of $(0.9, 0.999)$, gradient clipping at 1.0, and batch size of $32$. We continue the pretraining for $800k$ iterations. 

\subsection{Exploratory Data Analysis}\label{ssect:eda}
\begin{figure*}[h]
  \centering
  \includegraphics[width=1\textwidth]{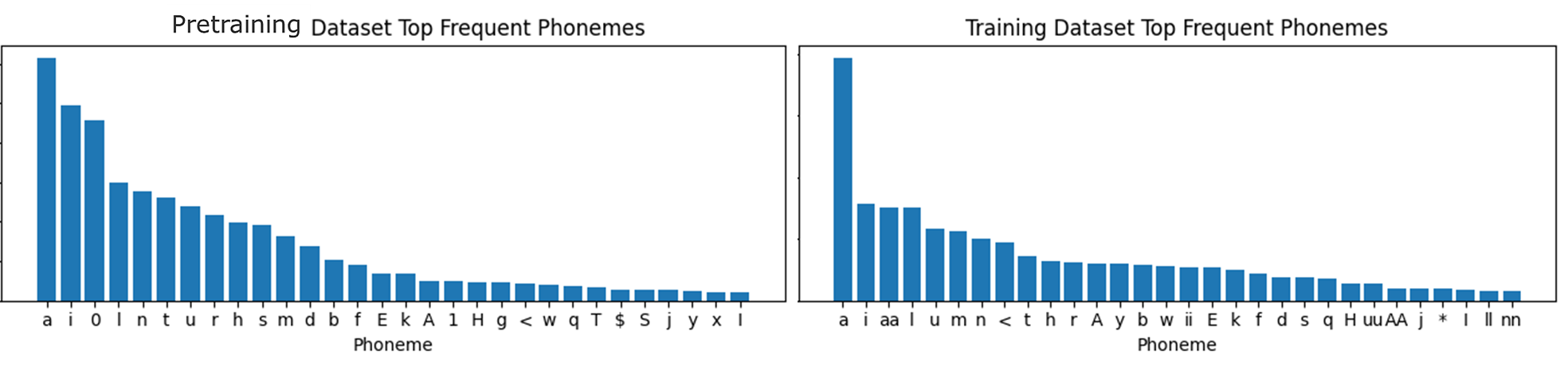}
  \caption{Histogram of top frequent phonemes in pseudo-labelled pretraining  and training datasets}
  \label{fig:phoneme-hist}
\end{figure*}

We have had a hypothesis that there is a discrepancy between pretraining data and fine-tuning one.
So, we plotted the histogram of the most frequent phonemes in both the pretraining and training datasets.
As shown in figure~\ref{fig:phoneme-hist}, the distributions of phonemes differ notably, particularly for elongated phonemes such as “aa,” “ii,” “uu,” and “AA.”.
This observation confirms the correctness of our hypothesis and highlights the importance of further fine-tuning on downstream task.

Prior to fine-tuning, we notice a difference between the phoneme inventory in the training dataset and the phonemes produced by the MSA phonetizer.
We align the phonemes as shown in Table~\ref{tab:phoneme_mapping}.
\begin{table}[h]
\centering
\begin{tabular}{lc}
Phonetiser Phoneme & Inventory Phoneme \\
\hline
II0 & II \\
I0I0 & II \\
I0 & I \\
I1 & I \\
ii0 & ii \\
i0i0 & ii \\
i0 & i \\
i1 & i \\
UU0 & UU \\
U0 & U \\
U1 & U \\
uu0 & uu \\
u0u0 & uu \\
uu1 & uu \\
u0 & u \\
u1 & u \\
\end{tabular}
\caption{Mapping from MSA phonetizer output to the training dataset phoneme inventory.}
\label{tab:phoneme_mapping}
\end{table}

\subsection{Fine-tuning}\label{ssect:ft}
After continuing pretraining, we performed vanilla fine-tuning for  the model on our ``Tuning dataset''~\ref{sssect:tuning-data}.
We used the same training parameters as that of continue pretraining.
\subsubsection{Tuning dataset}\label{sssect:tuning-data}
To further align the model with the task, we used the training set provided with the task, and created synthetic dataset to increase overall data size.
Preparing the synthetic data has went through two main stages: prepare the noisy text and generate corresponding audio files.

\paragraph{Prepare Noisy Text:} We downloaded the text of the holy quran and perturbed the text with what we consider to be valid noise.
The algorithm to generate valid noise is shown in algorithm\ref{alg:noise-generator}.
\begin{algorithm}
\small
  \caption{Noising Algorithm}
  \label{alg:noise-generator}
  \begin{algorithmic}[1]
    \Procedure{GenerateNoisyText}{text, arabic\_chars, noise\_map, max\_noise}
      \State $target\_noise \gets RandInt(1, max\_noise)$
      \State $new \gets$ empty list; $count \gets 0$
      \For{ch in text}
        \If{$count >= target\_noise$}
            \State Append ch
        \ElsIf{$UniformRandom(0,1) < p_{noise}$}
          \State $count \gets count+1$
          \State Choose noise type: delete / substitute / insert
          \If{substitute} \State Append RandChoice (noise\_map[ch])
          \ElsIf{insert} \State Append RandChoice(arabic\_chars), ch
          \EndIf
        \Else
          \State Append ch
        \EndIf
      \EndFor
      \State \textbf{return} Join($new$)
    \EndProcedure
  \end{algorithmic}
\end{algorithm}

\paragraph{Audio Generation:} We downloaded many audio files for various speakers to ensure the variety of data and to avoid overfiting over small set of speakers.
Then, we generated speaker embeddings using embedder module in XTTS-v2~\cite{casanova2024xtts}.
Finally, we converted the noisy text to audio files using XTTS-v2.

The resulted dataset is $60$ hours of audio files, and represented $30\%$ of Tuning data.

While selecting checkpoint for testing, we noticed a shift in distribution between our valid set and competition's test set.
Hence, we selected checkpoint saved after $2.5$ epochs for submission to balance generalizability and good performance on the downstream task.


\section{Results}
In this section, we illustrate the metrics used (section~\ref{ssect:metrics}), report quantitative results (section~\ref{ssect:quantitative}), and shows some examples from our qualitative analysis (section~\ref{ssect:qualitative}).

\subsection{Metrics}\label{ssect:metrics}
The system is evaluated using several complementary metrics. First, the \textbf{Correct Rate} measures the proportion of phonemes that are detected correctly, and is defined as \(1 - \text{Phoneme Error Rate (PER)}\). In addition, \textbf{Accuracy} captures the proportion of phonemes classified correctly as either pronounced correctly or mispronounced. To further distinguish system behavior, \textbf{True Acceptance (TA)} refers to cases where a correct phoneme is correctly accepted, while \textbf{True Rejection (TR)} corresponds to mispronounced phonemes that are correctly flagged. Conversely, errors are represented by \textbf{False Acceptance (FA)}, when a mispronunciation is missed, and \textbf{False Rejection (FR)}, when a correct phoneme is wrongly flagged. Beyond detection, \textbf{Correct Diagnosis (CD)} evaluates how often the system not only detects a mispronunciation but also identifies the specific mispronounced phoneme. Finally, the system’s classification quality is summarized through \textbf{Precision}, defined as \(\frac{TR}{TR + FR}\), \textbf{Recall}, defined as \(\frac{TR}{TR + FA}\), and their harmonic mean, the \textbf{F1-score}, computed as \(\frac{2 \cdot \text{Precision} \cdot \text{Recall}}{\text{Precision} + \text{Recall}}\).

\subsection{Quantative Analysis}\label{ssect:quantitative}
Table~\ref{tab:results} shows the results of our system under different setups: after adaptive continue pretraining, fine-tuning on the official training data of the task, and after fine-tuning on our Tuning data.
The results demonstrate that fine-tuning is essential for optimizing the system’s alignment with Qur’anic recitation assessment.
More importantly, they show the effectiveness of our synthetic data generation pipeline, achieving top performance across all of our systems.

\begin{table*}[h]
\centering
\small
\begin{tabular}{lccccccccc}
\hline
\textbf{System} & \textbf{F1$\uparrow$} & \textbf{Prec.$\uparrow$} & \textbf{Rec.$\uparrow$} & \textbf{CR$\uparrow$} & \textbf{Acc.$\uparrow$} & \textbf{TA$\uparrow$} & \textbf{FR$\downarrow$} & \textbf{FA$\downarrow$} & \textbf{CD$\uparrow$} \\
\hline
pretraining only & 0.1923 & 0.1091 & 0.807 & 0.5156 & 0.5117 & 0.5264 & 0.4736 & \textbf{0.193} & 0.4363 \\
fine-tuning \\
\hspace{3mm} training data & 0.4561 & 0.3327 & 0.7252 & 0.8714 & 0.8576 & 0.8954 & 0.1046 & 0.2748 & 0.568 \\
\hspace{3mm} Tuning data & \textbf{0.4726} & 0.3713 & 0.6501 & \textbf{0.8985} & \textbf{0.8701} & \textbf{0.9209} & \textbf{0.0791} & 0.3499 & \textbf{0.6873} \\
\hline
\end{tabular}
\caption{Performance on the Iqra’Eval 2025 leaderboard. CR = Correct Rate, Acc. = Accuracy, TA = True Acceptance, FR = False Rejection, FA = False Acceptance, CD = Correct Diagnosis.}
\label{tab:results}
\end{table*}

\subsection{Qualitative Analysis}\label{ssect:qualitative}
Table~\ref{tab:qual-results} presents examples from both the fine-tuning  on training set only setup and the continued pretraining-one.
Because of time constraints and high similarity between fine-tuning on training set only and on Tuning set, we leave its qualitative analysis for future work.
The results indicate that the system trained with pretraining alone fails to accurately predict phonemes associated with diacritics, particularly the ``shadda''.
This limitation is likely due to the rarity of such phonemes in the pretraining data as discussed in subsection \ref{ssect:acpt}.
This further confirms that adaptive continue pretraining was not sufficient and that we need for fine-tuning on the training set of the task.
\begin{table*}[h]
\small
\centering
\begin{tabular}{|p{3cm}|p{3cm}|p{4cm}|p{4cm}|}
\hline
\textbf{Ref. Aya Segment} &
\textbf{Recited Aya Segment (With Error)} &
\textbf{Pretrained System Output} &
\textbf{Fine-tuned System Output} \\
\hline

\begin{arabtext}يُغَشِّيكُمُ النُّعَاسَ أَمَنَةً مِنْهُ\end{arabtext} &
\begin{arabtext}يُخَشِّيكُمُ النُّعَاسُ أَمَنَةً مِنْهُ\end{arabtext} &
\texttt{ii x \$ ii k m l n E aa s m n h m n h} &
\texttt{y u x A \$\$ ii k u m u l nn u E aa s u \textless{} a m a n a t a n mm i n h u} \\
\hline

\begin{arabtext}إِيَّاكَ نَعْبُدُ\end{arabtext} &
\begin{arabtext}إِيَّاكَ نَعْبَدُ\end{arabtext} &
\texttt{y aa k n E b d} &
\texttt{\textless{} ii y aa k a n a E b a d u} \\
\hline

\begin{arabtext}ذَٰلِكَ الْكِتَابُ لَا رَيْبَ ۛ فِيهِ\end{arabtext} &
\begin{arabtext}ذَٰلِكَ الْكِتَابُ لَا رَيْبُ ۛ فِيهِ\end{arabtext} &
\texttt{* aa l i k l k t aa b l aa r ii b f ii h} &
\texttt{* aa l i k a l k i t a b u l aa r a y b a f ii h i} \\
\hline

\begin{arabtext}ٱلرَّحۡمَٰنْ\end{arabtext} &
\begin{arabtext}ٱلرَّرَّرَّحۡمَٰنْ\end{arabtext} &
\texttt{a l r H m n} &
\texttt{l rr rr rr a H m a n i} \\
\hline
\end{tabular}
\caption{Comparison Between Only Pretrained and Fine-tuned System}
\label{tab:qual-results}
\end{table*}

\section{Conclusion}

In this work, we illustrated our recipe to adapt Wav2Vec-BERT 2.0 on speech-to-phoneme task.
First, adaptively continued pretraining it on Arabic speech-phonemes corpora.
Second, we prepared synthetic data for fine-tuning phase by generating noisy text, convert it to phonemes using MSA-phonetizer, and generate corresponding speech for many speakers using XTTS-v2.
Our model scored first on IqraEval 2025, illustrating the ffectiveness of our approach.

\section{Acknowledgment}
We thank the IqraEval organizers\cite{elkheir2025iqraeval} for their support and for providing clear documentation and tools such as the evaluation API. We also acknowledge the contributors of the open-source datasets we used. This effort was supported by Applied Innovation Center (AIC) ‘s High Performance Computing facilities

\bibliography{references}

\end{document}